# Adversarial Attack Against Image-Based Localization Neural Networks

*Meir Brand, Itay Naeh, Daniel Teitelman*
Rafael - Advanced Defense Systems Ltd., Israel
br.meir@gmail.com, itay@naeh.us, dtyytlman@gmail.com

## Abstract

In this paper, we present a proof of concept for adversarially attacking the image-based localization module of an autonomous vehicle. This attack aims to cause the vehicle to perform wrong navigational decisions and prevent it from reaching a desired predefined destination in a simulated urban environment. A database of rendered images allowed us to train a deep neural network that performs a localization task and implement, develop and assess the adversarial pattern. Our tests show that using this adversarial attack we can prevent the vehicle from turning at a given intersection. This is done by manipulating the vehicle's navigational module to falsely estimate its current position and thus fail to initialize the turning procedure until the vehicle misses the last opportunity to perform a safe turn in a given intersection.

## Introduction

The future of transportation will be autonomous [2]. Self-driving cars [4] and drones [3] are already on the ground and in the air. These platforms rely on multiple sensors such as LiDAR [5,6,7], GPS [8], cameras [9,10], and additional IMUs [11] for estimating the state of the observed environment. Incorporating additional algorithmic approaches, such as sensor fusion [12,13,14,15], one can determine more accurately the action a given platform should take at each step of the way in order to reach its goal destination. In urban areas, GPS accuracy may be hindered by a high density of buildings [16]. In such cases, autonomous platforms can use image-based localization for self-positioning. This type of localization is done by taking a single image from the usually forward-looking camera of the vehicle and passing it through a neural network that was trained for providing the platform position and orientation, in the environment the network was trained upon.

Adversarial attacks are carefully crafted patterns that disrupt a neural network output when introduced as an input [17, 18, 19]. In this work we have crafted a patch that was implemented on a street billboard placed in front of a traffic intersection. The patch pattern disrupts the navigation system of the autonomous vehicle and prevents it from reaching its desired destination. Exposing such vulnerabilities in deep neural networks in general is beneficial to industry and academic researchers, in order to increase awareness of this matter. Figure 1 shows the top-view of the simulated urban environment (a), navigation path (ABC) (b), and a single image taken by the car along the driving route (c). This simulated environment was modeled with Brushify-Urban Buildings Pack [20] on the Unreal Engine 4 [1].

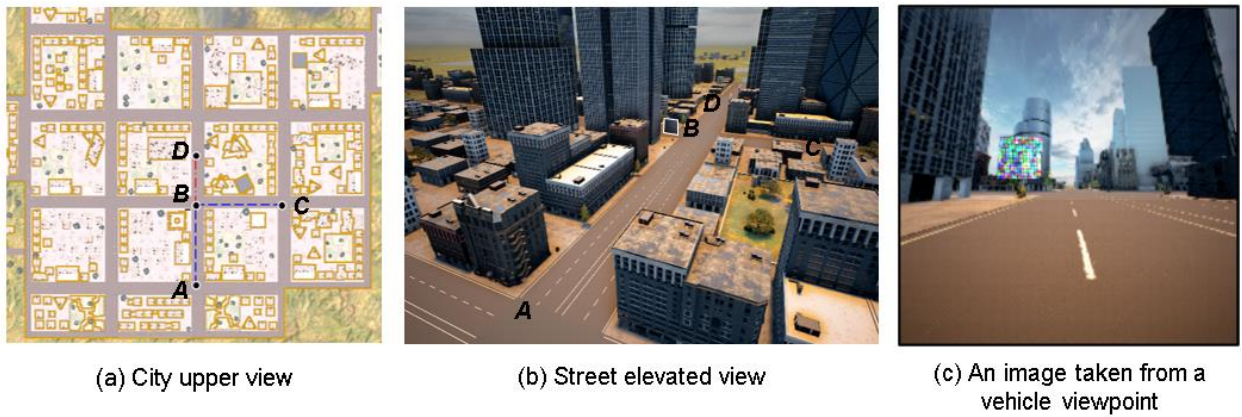

Figure 1 - The simulated urban environment

# Related Work

The field of adversarial attacks against neural networks for autonomous vehicles spans over a wide range of research areas, and in this section we briefly review the most widely-known works. In his pioneering work I. Goodfellow et. al. [17] described a way to generate adversarial examples, followed by N. Carlini et. al. [22]. In 2017 a group of researchers led by T. B. Brown [23] has shown how a localized adversarial patch in the real world can hinder the prediction of a neural network completely, while encompassing only a small part of the image. From 2020 two additional important papers were published, which both continue the leap into making adversarial attacks applicable to the real world, in the first one by Z. Kong [24] it was shown that an advertising sign can be used as an effective adversarial patch for influencing the steering module of an autonomous vehicle, and a work by H. Salman [25] described the existence of unadversarial examples, not just for 2D objects but for 3D as well. The described patch is robust to different lighting conditions, orientations and camera views. To the best of our knowledge, attacks of a self localization neural network were not published before.

# Research method

This proof of concept aims to demonstrate how an adversarial patch on a street billboard can induce wrong navigational decisions. Real world autonomous platforms are varied and diverse, so in order to approach this abundance we will define a target platform which consists of an autonomous car with two relevant modules. The first is the navigation module, which dictates the general path that the platform should take from one point to another and whether to turn or not at each junction. The second module is the automatic driver, which is a tactical module that determines how to handle the car's immediate actions, keeping to the lane, turning at the proper arc, and not causing any accidents with the surrounding actors and environment. It would be helpful to consider the navigation module as the master system who orders the automatic driver where to turn.

Within the scope of this work only the navigation module will be discussed since the attack will be performed before the automatic driver will be engaged for turning.

In our scenario the navigation plan of the car is to drive along a path ABC shown in Fig 1.a, with the help of the navigation module. The adversarial attack on the navigation module will cause the car to miss the right turn in intersection B and keep driving straight to point D instead of driving to point C.

The main components needed for performing our adversarial research were:
a. Simulating an urban environment and controlling it by an external software (python) to produce a database of street images.
b. Training a localization CNN which can estimate vehicle position and orientation {x, y, direction} with sufficient accuracy based on a single image taken from the vehicle's front camera.
c. Crafting an adversarial patch in order to produce an adversarial attack on the localization CNN.
d. Handling the navigation system, which is in charge of reporting to the vehicle self-driving system its basic commands ("continue straight" or "intersection ahead turn right", etc.).

## a) The simulated urban environment

There are a variety of relevant Unreal Engine environments for this kind of demonstration (City, Neighborhood, Urban etc.). The main parameters we chose for this work were: an environment that simulates city streets which are not too dense, the environment has about 10 streets with few intersections between them, and the buildings are diverse. The streets have sidewalks with some trees and plants along them and driving lanes for cars. We have found that the Brushify-Urban Buildings Pack [21] was suitable for this research. The city map (upper view of the city) is shown in Fig 1.a. The yellow border (squares) in this figure are the buildings contours. The variety of buildings which include shape, texture, color and height can be seen in Fig 1.b. The typical width of a traffic road in this environment is about 50 [m] (4 lanes) which made us choose a wide field of view (FOV) of 120° for the vehicle's camera. This setup was used for the images dataset taken for the training of the localization CNN, for the adversarial patch development and for the vehicle path navigation. Brushify-Urban Buildings Pack has further entities like lakes, open fields and peripheral suburbs outside the city zone shown in Fig 1.a. For this research we have limited our zone of interest to be the streets shown in this figure. This zone is about 1.4 [km] length and 1.1 [km] width.

## b) Localization Convolutional Neural Network (CNN)

The input/output of our localization CNN was selected to be:
- Input image size of 224x224x3 RGB pixels.
- Output vector consists of 3 parameters: x coordinate, y coordinate and θ which is the angle the car is facing.
- All images were taken about 0.4÷1.0 [m] above the ground in order to represent images taken from a car's front view camera. Also, 7° [deg] of elevation was given to the camera to include more informative features in the FOV.

A dataset of 100K simulated images facing the four major directions was taken on the streets for the training and testing (90%-10%, respectively) phases of different CNN models. Images facing outward from the city toward open territories were not taken.

Few different CNN models were trained and tested. The architecture of the final CNN model was with 1,060K parameters and it is presented in figure 2.

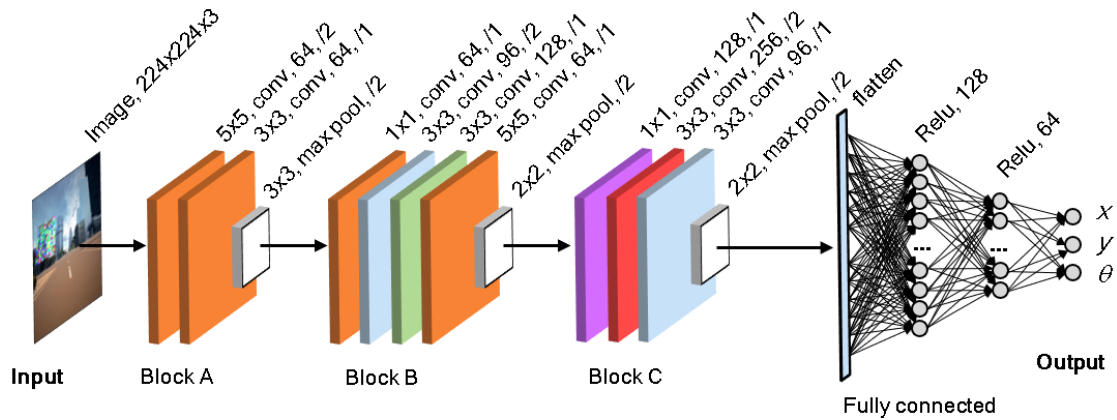

Figure 2 - The architecture of the localization CNN model. All convolutional layers were Relu activated. In this figure, "5x5, conv, 64, /2", the first conv layer was with 64 filters of 5x5 kernel and 2 pixels strides.

Due to the fact that the angular accuracy of the models was much better than the positioning accuracy, we rescaled the loss terms between the position and the direction. The trained CNN model had a mean positional error of 28.6 [m] on the training set and 30.4 [m] on the test set. Also, the model mean angular error was 6.2° for the training set and 6.6° for the test set. The position and direction error distribution of the model is presented in figure 3.

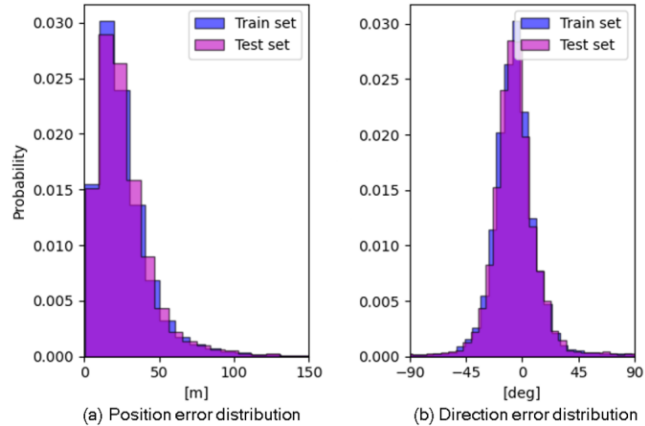

Figure 3 - Probability distributions of train and test sets position and direction errors

The images (a) and (b) in figure 4 show test set errors of the trained CNN model. Figure 4.a shows true positions and directions with blue lines and the CNN prediction positions and directions with magenta lines. Figure 4.b shows the same data and results with focus on position deviation. In figure 4.b the dots are test images true positions. The dots color represent the deviation errors according to a color scale (0÷60 [m] color bar). Also, a colored line connects the true positions, where the test image was taken, to their CNN predicted positions. The dotted black rectangular in figure 4.b is the AB street where the adversarial attack was demonstrated.

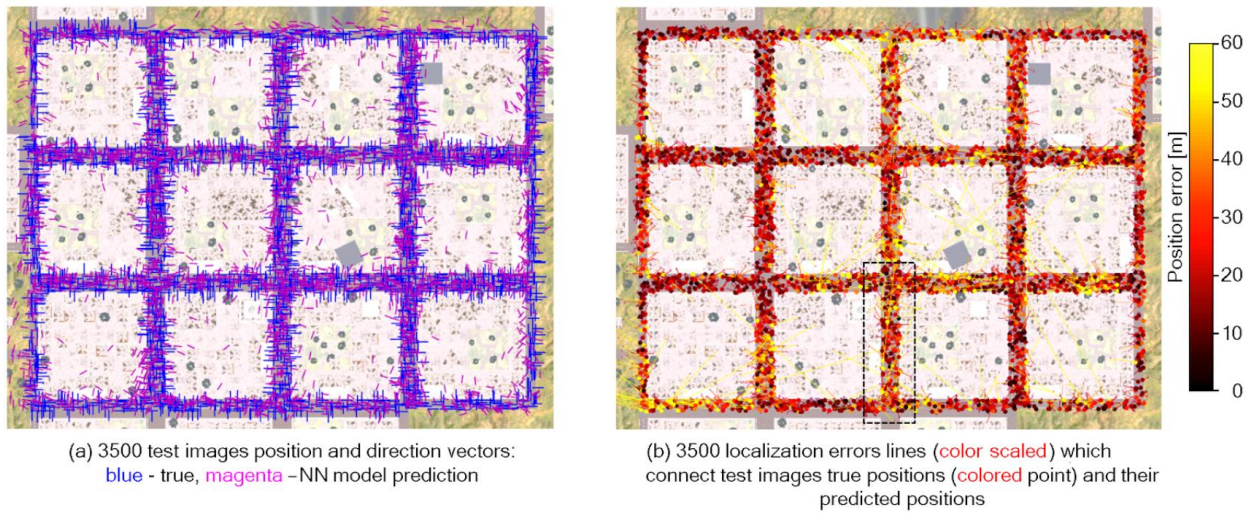

(a) 3500 test images position and direction vectors:
blue - true, magenta –NN model prediction

(b) 3500 localization errors lines (color scaled) which connect test images true positions (colored point) and their predicted positions

Figure 4 - Position and direction errors of 3500 test samples presented on the city map

From the above figure we can learn that the trained CNN errors are relatively spatially homogeneous and the mean position error is about ~25 [m], which is well within half width of a street.
In order to analyze the neural network model behavior, we investigated the contribution of each area in an input image to the change in the loss value. Higher contribution values, with a kernel of 4x4 pixels, show which area of the image is more dominant in the localization task. Three street images with their corresponding per-pixel loss matrix are shown in figure 5.

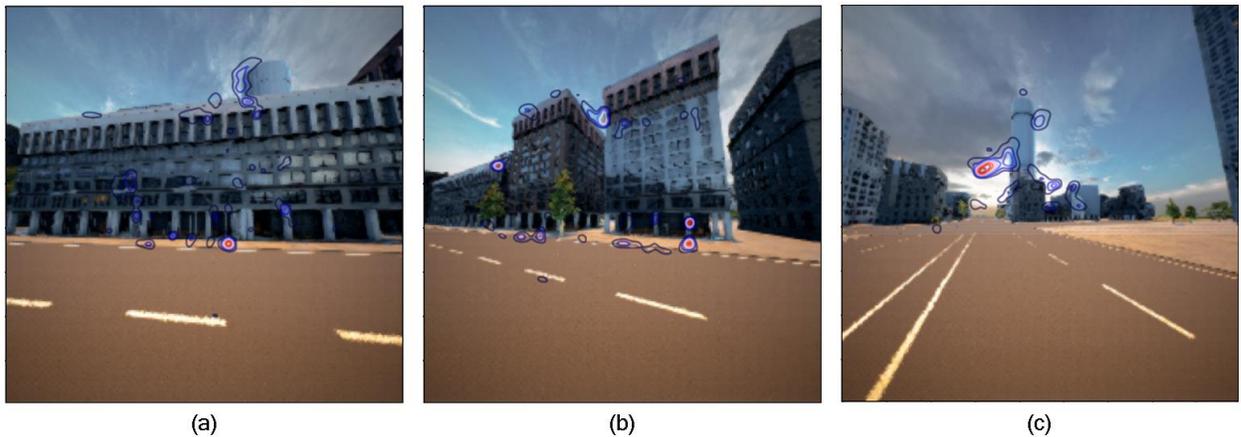

(a)  (b)  (c)

Figure 5 - Influence contours are presented on top of their input images

The figures a,b,c show that the loss function is focused on features such as building contours, sidewalks borders, windows and building columns. The influence contours in figure 5.c include some sky features as an additional source for its prediction. We found that the CNN model categorically prefers building and street features over sky features. Also, the sky features can assist the CNN model to learn directions but not location, due to the "sky-dome" rendering process.

## c) Adversarial patch development algorithm

There are several possible ways for applying an adversarial patch, and in this work we chose a street billboard positioned on the sidewalk before the intersection. In order to achieve robust and consistent localization attacks we have considered two approaches:
a) Constant position error - the CNN model will predict a far constant position not near the driving path and street (like point C in figure 1.b).
b) Backward or forward relative position shift - the CNN model will predict a position shift that is more forward (in the direction of B-C) or more backward (in the direction of B-A) relative to the vehicle's current true position.

For this work we selected the backward relative position shift (b). A value function was crafted to assess the patch influence on a set of the vehicle's street images. Backward shift means that the predicted position will be behind the vehicle's true position with respect to the vehicle's driving direction. Therefore, in the desired case, the vehicle's navigation system will not have enough time to prepare for turning while the vehicle itself is passing the "Last Turning Point" of the intersection, as will be explained later.

Considering the relatively large scale of the urban model, and the problems it imposes on packages like Pytorch3D or other gradient-able rendering pipelines with our available hardware, we have used a black box setup in order to craft the adversarial patch. With this approach, the algorithm suggests a change in the adversarial pattern, assesses the proposed new pattern with respect to the desired adversarial loss, and adopts the change if it was improved.

In this research we chose 16x16 colored pixel resolution for the adversarial billboard patch (see figure 6.a.). Figure 6.b shows a sample of a vehicle's street image taken at the position shown in figure 6.c by a blue dot. The evaluation set of the adversarial effect of the optimization process of the patch consists of several positions along the street from point A to intersection B.

Initially, we used the Unreal Engine as part of the algorithm tools for automatically implementing a patch on the street billboard and then taking a set of images along the street. This concept was found time consuming and we chose a bypass approach. We calibrated a conversion table that enabled us to directly add a colored patch (pixel by pixel) on a street image taken with a black billboard in it. The billboard size, position and shape is different in each image, therefore the conversion table was unique for each vehicle's street image. So, the algorithm activated this routine for each image in the training set instead of using the Unreal engine for rendering the adversarial patch.

In order to improve the black-box design algorithm efficiency we have included a few optimization concepts. The algorithm was iterative, a value function was defined to assess patch influence on a set of street images and changes in the patch were moderate (small RGB changes and small spatial area of the patch) relative to the previous good patch. Also, if the algorithm did not succeed after several iterations to find a better patch, a bad perturbation was adopted as a good one.

When using the value function of L2 loss, a small number of samples taken along the street were significantly improving while all the others were not. For this reason we included a fairness based value function. The formulation of which is:

$$V(\Delta) = \frac{1}{(n_B + n_F)^2} \cdot \left( n_B \cdot \sum_{j=1}^{n_B} |\Delta_j^B|^\alpha - n_F \cdot \sum_{j=1}^{n_F} |\Delta_j^F|^\alpha \right), \quad 0 < \alpha < 1$$

where: $n_F$ and $n_B$ are the number of samples that their predicted location moved forward or backward, respectively, due to the patch. $\Delta_j^F$ and $\Delta_j^B$ are the distances between the predicted and true positions of sample j, of groups $n_F$ and $n_B$, respectively. $\alpha$ is the regularizing power of the distance. This kind of value function improves the generalization and reduces the effect of extreme singular samples.

The patch presented in figure 6.a was designed with the above algorithm and was used to attack the localization neural network before reaching intersection B. The predicted position of the CNN model is presented with a gray x in figure 6.c. The blue dot in figure 6.c is the true position where the vehicle's street image (figure 6.b) was taken. The magenta star in figure 6.c is the estimated vehicle position and it is explained in the next paragraph.

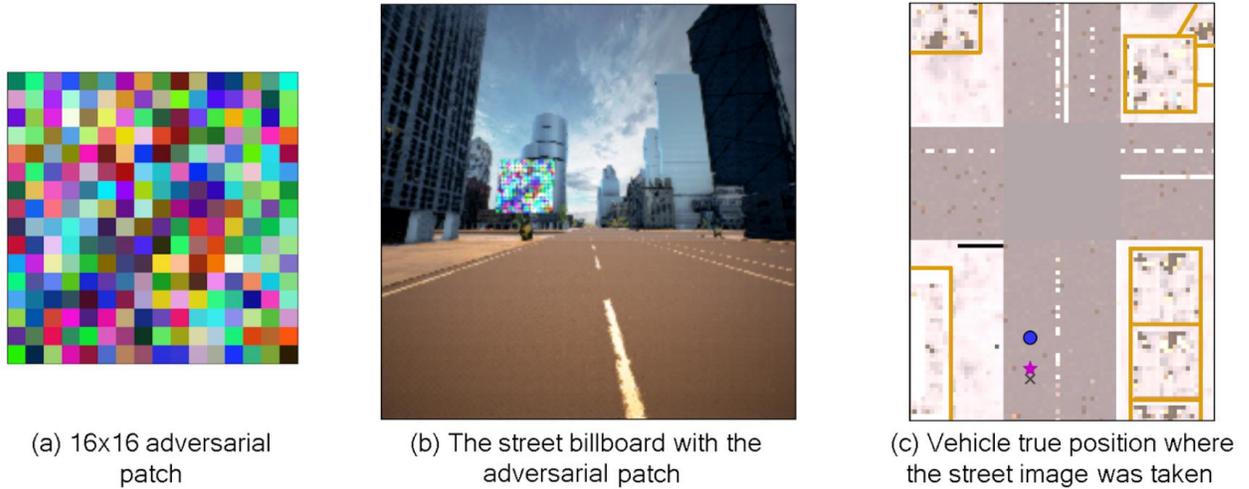

(a) 16x16 adversarial patch    (b) The street billboard with the adversarial patch    (c) Vehicle true position where the street image was taken

Figure 6 - The adversarial patch (a), vehicle's street image with the patch on the street billboard (b) and true and predicted positions on the city map (c).

### d) Vehicle navigation system concept

As stated earlier, the autonomous car concept for the adversarial demonstration has two separate systems: the navigation system and the self-driving system. The vehicle navigation system is based on low-frequency positioning data based on the position predictions of the CNN model received from immediate street images. In order to reduce positioning errors, a linear regression was applied with the last seven predicted positions and the current one to produce a self-position estimation $\{x_j^E, y_j^E\}$ of the current vehicle position. The estimated position shown in figure 6.b is presented with a magenta star in figure 6.c.

The navigation system is in charge of providing the driving module an indication for turning an adequate amount of time before reaching the intersection. An exemplary turning protocol may consist of slowing down the vehicle, increasing the detector's sampling rate, expanding the processed field of view, and other required actions for a safe turning. The region these actions take place starts with the green line $L_1$ in figure 7.b and ends with the red line $L_2$. If the vehicle's self-driving system is informed that it is approaching an intersection and the vehicle's true position is between these two lines, then the vehicle sensors and algorithms will succeed in performing a safe turn. But, if the vehicle's true position is too close to the intersection (passed the red line) and the localization module did not provide an indication that the vehicle is approaching the street intersection for turning, so the turning protocol would not be able to execute and perform a safe turn, thus continuing straight.

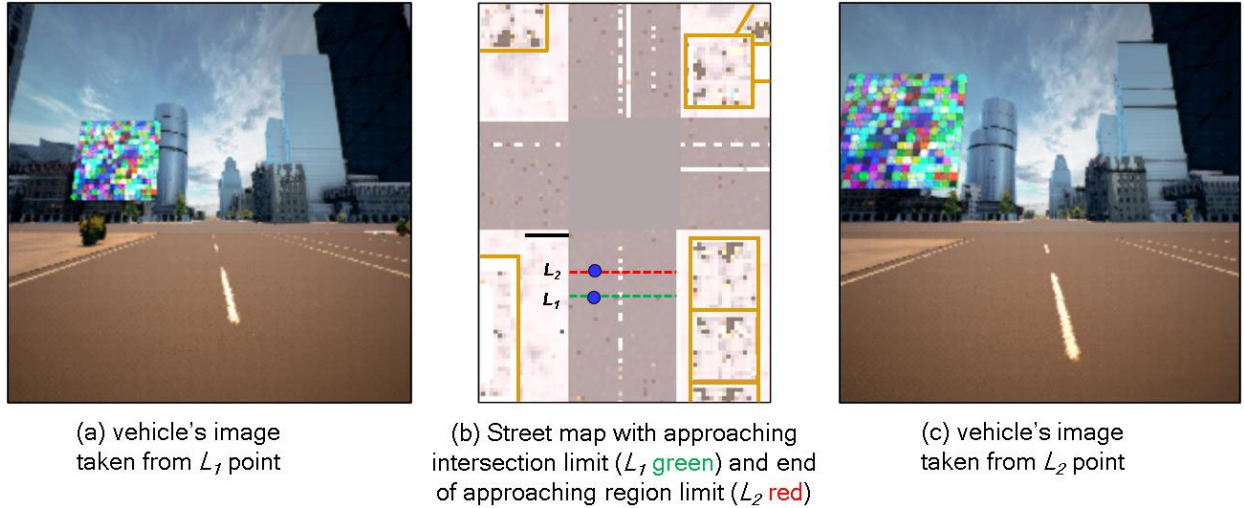

(a) vehicle's image taken from $L_1$ point

(b) Street map with approaching intersection limit ($L_1$ green) and end of approaching region limit ($L_2$ red)

(c) vehicle's image taken from $L_2$ point

Figure 7 - Approaching intersection region's limits

Therefore, if the adversarial patch can affect the localization system and delay (backward shift) its position estimation before $L_1$ while the true position will be in front line $L_2$ closer to the street intersection, the turning will not be executed and the attacker's goal was achieved.

# Results

The adversarial patch was designed to achieve a significant backward localization shift from the vehicle's actual position. This situation delays the report from the navigation module to the driving module, thus preventing the turning at the intersection. The adversarial patch that achieved this goal is presented in figure 7.a and 7.c.

The graph in figure 9.a presents the adversarial attack relative effect on the *y* distance to the intersection B. For example, the image shown in figure 8.b was taken around -80 [m] before the intersection B and its exact position can be seen as the small blue dot on the bottom of figure 8.a.

In addition, the predicted position ($y_j^P$) and the expected position ($y_j^E$) of image *j* are presented in figure 9 as relative distance from intersection B. The *x* axis in figure 9 graphs is the relative true position ($y_j^T$) relative to intersection B. The relative true positions of the 40 images of the test set are shown as linear blue dots 'o' in figure 9.a. The relative CNN predicted positions based on these images are shown by the gray 'x' mark in the figures. If the 'x' mark is above the image's true position 'o' then the prediction is ahead of the actual vehicle's position otherwise the prediction is behind (backward shift). The navigation system's estimation of the position of the vehicle is shown with a magenta star '☆'. The vehicle estimation module depends on the current predicted position and previous 7 positions CNN predictions.

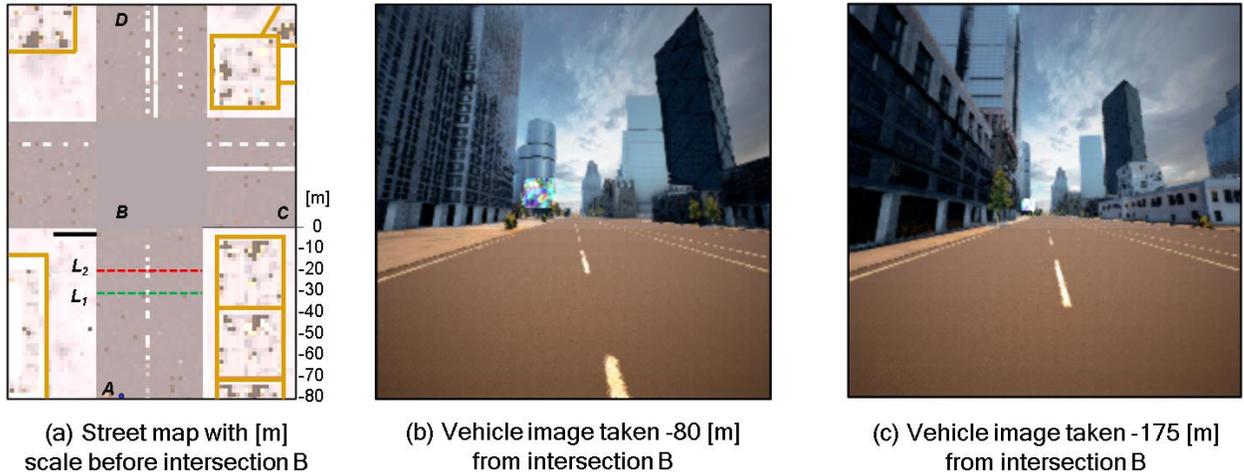

(a) Street map with [m] scale before intersection B

(b) Vehicle image taken -80 [m] from intersection B

(c) Vehicle image taken -175 [m] from intersection B

Figure 8 - Street map with scaling and images taken around -80 [m] and -175 [m] from intersection B

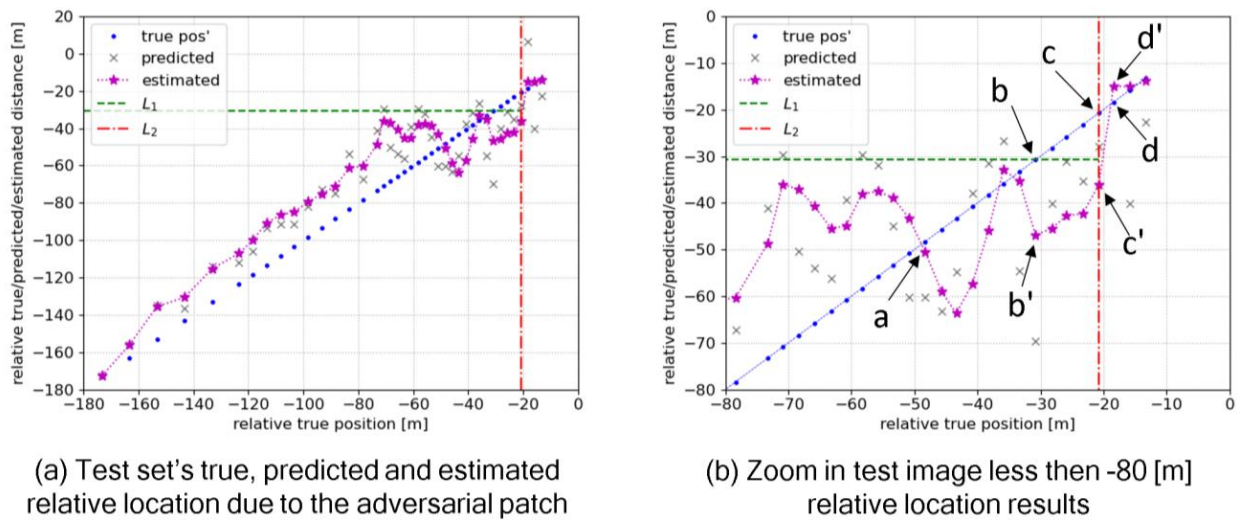

(a) Test set's true, predicted and estimated relative location due to the adversarial patch

(b) Zoom in test image less then -80 [m] relative location results

Figure 9 - Adversarial patch position influence on the test set vehicle's images

Almost all the vehicle's predicted and estimated positions based on images that were taken in 50 [m] and less from the intersection B showed significant backward shift due to the adversarial patch. From around point (a) in figure 9.b the influence of the patch is noticeable and the true position of the vehicle (blue dots) are higher than the prediction. Point (b) is when the vehicle's true position is passing the $L_1$ line (green line) and entering the "approaching intersection B" region. Since the estimated position of the vehicle at the same moment, point (b'), is more than ~17 [m] backward, the localization system is not aware that the vehicle has entered the intersection B approach region and will not provide any indication to the driving module. Point (c) is the moment when the vehicle's true position passed the end of approach line $L_2$ (red line). At this moment the vehicle's estimated position (c') still had not entered the intersection B approach region by passing $L_1$ line. Since the localization system did not report that the vehicle is approaching intersection B to the self-driving system, the turning will not be executed, and the vehicle will continue straight towards point D (see figure 8.a).

We have checked this conceptual localization attack with 3 more patches: black patch, white patch and a random patch. Figure 10 shows the effect of those three patches with the adversarial patch on the same plot as figure 9.b.

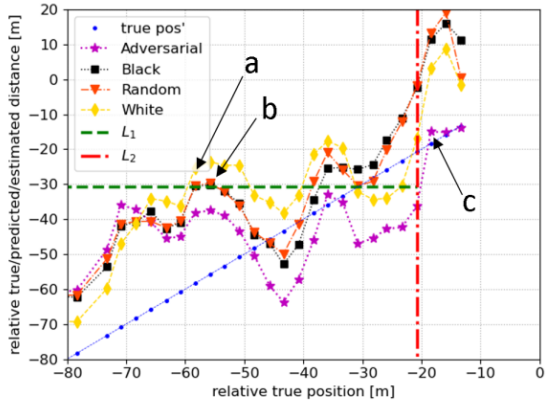
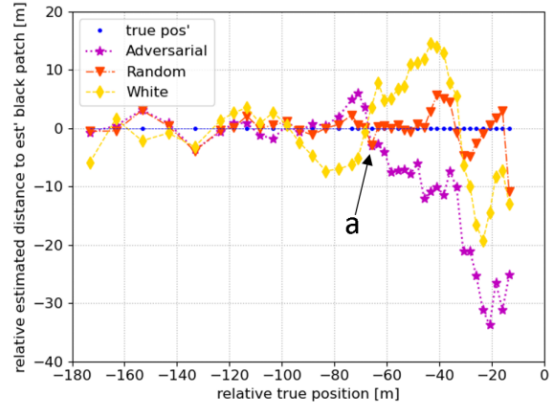

(a) Adversarial, black, random and white patches estimated relative locations

(b) Estimated locations of the adversarial, random and white patches relative to the black patch results

Figure 10 - Black, random and white patches localization effects

Points (a), (b) and (c) shown in figure 10.a are the first time that the estimated positions of the white, black, random and adversarial patches entered the intersection B approach region (line $L_1$). Point (c) relates to the crafted adversarial patch and it is the first time that the vehicle's CNN model's estimated position crosses the $L_1$ line (approaching intersection B region) but also crosses the $L_2$ line (end of approaching intersection region). This means that the vehicle navigation system didn't recognize that the vehicle was in the intersection approach region until it actually passed its last chance to turn. The black, white and random patches don't show any influence on the vehicle navigation system since it recognizes the intersection approach region early and prepares the vehicle to take the turn.

Figure 10.b shows the estimated localizations of the vehicle while the street billboard has a black, random and white patches separately, but in this figure we calculated the estimated results relative to the black patch. Point (a) in this figure is the first time that the adversarial position estimation is backward to the black patch estimations. This starts -60 [m] before intersection B and reaches a maximum backward shift of -33 [m] (along the y axis) relative to the black patch position estimation.

# Conclusions

In this work, we have demonstrated the ability to adversarially fool an autonomous vehicle's localization neural network by inducing a backward localization error by an adversarial patch. This prevented the initialization of the turning protocol in time to execute a turning, thus causing a navigation error to the vehicle by missing a turn. The billboard patch induced a substantial error and acted robustly to different on-road positions, generally inducing an error inversely proportional to the distance to it, preventing sudden repositioning. The adversarial patch was compared to blank, white and random patches in order to demonstrate its effectiveness.

______________________________________________